\documentclass[conference]{IEEEtran}
\IEEEoverridecommandlockouts

\usepackage{booktabs}
\usepackage{multirow}
\usepackage{array}
\usepackage{graphicx}
\usepackage{url}

\title{Cross-Dataset Stability of Expert-Informed Skill Prompting and Fine-Tuning for Chinese Metaphor Identification}

\author{
\IEEEauthorblockN{Yufeng Wu and Meichun Liu}
\IEEEauthorblockA{City University of Hong Kong}
\thanks{Corresponding author: Yufeng Wu.}
}

\begin{document}
\maketitle
\bstctlcite{BSTcontrol}

\begin{abstract}
Metaphor-identification performance can change markedly across datasets that differ in text distribution and annotation policy. We examine whether a fixed expert-informed procedure produces a more even cross-dataset profile than task-specific parameter adaptation. Four prespecified conditions are compared for Chinese sentence-level metaphor identification: BERT fine-tuning (BERT-FT), QLoRA-based large language model fine-tuning (LLM-FT), direct zero-shot LLM prompting (LLM-ZS), and zero-shot prompting with a frozen procedural Skill (Skill-ZS). The Skill operationalizes established criteria involving contextual meaning, basic meaning, contrast, and comparison. Evaluation covers CMRE Test and two external datasets, CCIME and CMC. Fine-tuned scores are means over three seeds, whereas each zero-shot score comes from one deterministic configuration. Fine-tuning remains strongest on the native test set: BERT-FT reaches 91.76 Macro-F1. LLM-FT has the highest external mean (83.52), while Skill-ZS is close at 82.92 and has both the highest external floor (82.64) and the smallest observed range across all three datasets (4.08 points). In the matched zero-shot comparison, adding the Skill reduces metaphorical predictions on every dataset. This sharply lowers false positives on CCIME but increases false negatives on CMRE Test and CMC. The results position expert-informed Skill prompting as a complementary route to more even observed cross-dataset performance, while fine-tuning retains its advantage in native-data accuracy. To our knowledge, this is the first study to compare an expert-informed procedural Skill with task-specific fine-tuning in the same cross-dataset evaluation of Chinese sentence-level metaphor identification.
\end{abstract}


\section{Introduction}

Metaphor-identification systems often change behavior when the evaluation dataset changes. Differences in genre, construction mix, class prevalence, and annotation policy can alter both overall performance and the balance between false positives and false negatives \cite{aghazadeh2022metaphors,uduehi2024expectation,reimann2024religious}. A high source-aligned score therefore captures only one part of the evaluation problem; the external performance floor and the variation across datasets are also consequential.

The task itself depends on deciding whether an expression has a contextual meaning that contrasts with, yet can be understood in relation to, a more basic meaning. Manual procedures such as MIP and MIPVU organize this judgment into explicit steps \cite{pragglejaz2007mip,steen2010mipvu}. Computational systems have represented related knowledge through model architecture, contextual interaction, explicit basic-meaning modules, prompts, and rule scripts \cite{mao2019sequential,choi2021melbert,li2023basic}. These alternatives place linguistic expertise in different parts of a system and may consequently produce different cross-dataset profiles.

Fine-tuning places task adaptation in parameters learned from labeled examples. An instruction Skill instead supplies an operational procedure at inference time while leaving the model parameters unchanged. We use \emph{Skill} to denote a reusable instruction artifact that specifies a multistep inference procedure rather than an in-context demonstration set or learned adapter. Its criteria may transfer across datasets differently from regularities learned from one training distribution.

We evaluate these routes under a common Chinese sentence-level formulation. BERT-FT and LLM-FT represent supervised parameter adaptation. Skill-ZS supplies the base LLM with a frozen expert-informed procedure derived from published annotation protocols and metaphor-identification research. Direct LLM-ZS uses the same base model and runtime without the Skill, providing a matched comparison of the procedure's aggregate effect.

We ask three questions. RQ1 asks how the four conditions differ in observed cross-dataset stability, summarized by native Macro-F1, external mean, external floor, the signed native--external mean gap, and three-dataset range. RQ2 examines the dataset- and class-specific performance trade-offs behind these summaries using accuracy, metaphor-class precision, recall, F1, and Macro-F1. RQ3 asks how adding the Skill redistributes aggregate decisions relative to the matched LLM-ZS condition. To our knowledge, this is the first study to compare an expert-informed procedural Skill with task-specific fine-tuning in the same cross-dataset evaluation of Chinese sentence-level metaphor identification \cite{reimann2025mipvu,huangliu2026rule,fuoli2026llm}. It further contributes a criterion-specific account of their complementary performance profiles and an aggregate analysis of the decision shift introduced by the Skill.

\section{Related Work}

\subsection{Expert Knowledge in Metaphor Identification}

MIP and MIPVU make metaphor decisions more reproducible by requiring analysts to establish contextual meaning, identify a more basic meaning, and assess contrast and comparison \cite{pragglejaz2007mip,steen2010mipvu}. Computational models have incorporated these principles in different forms. Sequential metaphor identification uses theory-inspired contextual signals, MelBERT models contextualized interaction, and explicit basic-meaning modeling represents the comparison directly \cite{mao2019sequential,choi2021melbert,li2023basic}. ContrastWSD further combines sense information with an MIP-based contrast test \cite{elzohbi2024contrastwsd}.

Generative models allow operational knowledge to be expressed in prompts rather than encoded in model architecture. Recent work evaluates zero-shot metaphor labels and rationales, MIPVU-inspired LLM procedures, and explicit rule scripts for Chinese metaphor identification \cite{chen2024reasonable,reimann2025mipvu,huangliu2026rule}. These studies motivate an expert-informed Skill. The present Skill adapts selected operational principles to sentence-level binary classification rather than implementing the full lexical-unit MIPVU procedure.

\subsection{Fine-Tuning, Prompting, and Dataset Change}

Metaphor-identification systems do not transfer uniformly across datasets. Cross-lingual research established that learned representations can support transfer across language-specific datasets \cite{tsvetkov2014crosslingual}. Later studies directly evaluated cross-dataset and cross-language generalization, including under multiple out-of-distribution settings, as well as cross-genre performance losses \cite{aghazadeh2022metaphors,uduehi2024expectation,reimann2024religious}. Broader NLP research likewise distinguishes source-aligned evaluation from performance under a changed target distribution \cite{yang2023ood}.

LLM-based studies add a second comparison: parameter adaptation versus direct or procedure-guided prompting. Recent studies examine surface-sensitive LLM metaphor judgments and multilingual evaluation of metaphor detection \cite{sanchezbayona2025surface,sanchezbayona2026meta4xnli}. Fuoli et al. compare retrieval-augmented generation (RAG), prompt engineering, and fine-tuning for full-text metaphor identification \cite{fuoli2026llm}. The present study focuses on Chinese sentence classification and asks whether explicit procedural guidance produces a more even observed cross-dataset profile. It combines a four-regime comparison with a matched direct-versus-Skill zero-shot intervention, allowing performance level and cross-dataset variation to be considered together.

\section{Methods}

We compare four ways of supplying task knowledge while holding the sentence-level label space and evaluation rows constant. CMRE Dev was used for supervised model selection, and both zero-shot conditions were finalized before formal evaluation. Implementation-level configurations and reproducibility records are provided in the supplementary package.

\subsection{Task, Data, and Overlap Control}

Each sentence receives either the metaphorical or non-metaphorical label. CMRE supplies the training, development, and native test partitions \cite{chen2023cmre}. We used its released sentence labels and retained simile relations labeled as positive. CCIME contributes the 1,200 publicly labeled Task~1 development rows \cite{ccime2026}, and CMC contributes its released held-out Chinese split \cite{li2022cmgen}. Only their binary metaphor labels were used. Table~\ref{tab:data} reports the retained data.

\begin{table}[t]
\centering
\caption{Dataset composition after CMRE Train/Dev overlap removal.}
\label{tab:data}
\footnotesize
\begin{tabular}{lrrr}
\toprule
Dataset & Total & Metaphorical & Non-metaphorical \\
\midrule
CMRE Train & 6,794 & 3,397 & 3,397 \\
CMRE Dev & 850 & 425 & 425 \\
CMRE Test & 850 & 425 & 425 \\
CCIME & 1,162 & 567 & 595 \\
CMC & 698 & 274 & 424 \\
\bottomrule
\end{tabular}
\end{table}

Before evaluation, exact and normalized-exact matches with CMRE Train or Dev were removed, excluding 38 CCIME rows and two CMC rows. Released duplicate rows within an evaluation view retained distinct identifiers so that the original row-level units were preserved. The resulting rows were shared by all four conditions. The annotation policies nonetheless differ: CMRE retains released-positive similes as metaphorical, whereas CCIME places similes and hyperbole in its non-metaphorical class. We therefore interpret CCIME as combining dataset shift with annotation-policy shift. Dataset versions, normalization rules, duplicate records, and overlap reports are documented in the supplement.

\subsection{Four Knowledge-Supply Regimes}

Table~\ref{tab:design} summarizes the four knowledge-supply regimes. BERT-FT and LLM-FT learn from CMRE labels; LLM-ZS and Skill-ZS receive no labeled examples. The four-way comparison describes practical systems with different architectures and adaptation costs. The zero-shot pair forms the controlled comparison: both use the same base model, runtime, prompt shell, output format, and evaluation rows, with the frozen Skill as the only condition-level addition.

\begin{table}[t]
\centering
\caption{Knowledge-supply regimes and formal run basis.}
\label{tab:design}
\footnotesize
\setlength{\tabcolsep}{3.2pt}
\begin{tabular}{lcccc}
\toprule
Condition & Labels & Updated & Skill & Runs \\
\midrule
BERT-FT & CMRE & full model & no & 3 \\
LLM-FT & CMRE & QLoRA & no & 3 \\
LLM-ZS & none & none & no & 1 \\
Skill-ZS & none & none & yes & 1 \\
\bottomrule
\end{tabular}
\end{table}

\subsection{Fine-Tuned Conditions}

BERT-FT used \texttt{hfl/chinese-roberta-\allowbreak wwm-ext-large} with a two-label sequence-classification head \cite{cui2020chinese}. The full model was optimized with AdamW at a learning rate of $2\times10^{-5}$, effective batch size 16, and maximum sequence length 512 for up to five epochs. Checkpoint selection maximized CMRE Dev Macro-F1, with lower Dev loss and then earlier epoch as tie-breakers. A fixed 0.5 threshold produced the labels.

LLM-FT adapted the text-only \texttt{Qwen/Qwen3.6-27B} model \cite{qwen2026qwen36} with 4-bit NF4 QLoRA \cite{hu2021lora,dettmers2023qlora}. Rank 8 and $\alpha=16$ yielded 23.76M trainable parameters. Assistant-only targets contained one JSON label, with loss restricted to target tokens. AdamW used a learning rate of $10^{-4}$, effective batch size 16, and maximum sequence length 256 for up to three epochs. Checkpoints were selected by the same CMRE Dev criterion as BERT-FT.

Both fine-tuned conditions used seeds 42, 43, and 44 and selected one checkpoint independently per seed before test prediction. The supplement reports the complete optimization, target-module, model-revision, and runtime configurations.

\subsection{Zero-Shot Control and Expert-Informed Skill}

LLM-ZS used the same Qwen base without parameter updates. Skill-ZS added the frozen Skill to this matched zero-shot configuration. We use \emph{zero-shot} to mean that no labeled CMRE, CCIME, or CMC example was used for parameter updates or supplied as an in-context demonstration. The direct prompt was dataset-agnostic and example-free.

The Skill translates established identification principles into six operations: establish sentence context, identify a candidate mapping, compare contextual and basic meanings, test contrast and comparison, reject unsupported cases, and aggregate the analysis to one sentence label. Its wording was refined through seven iterations on 60 AI-assisted synthetic specification sentences, balanced between the two labels. These items were used to revise the instruction rather than to estimate performance. They were excluded from formal evaluation, and no formal evaluation sentence, score, or error informed Skill development.

Both zero-shot conditions used deterministic decoding and the same constrained JSON label format. All 2,710 outputs per condition satisfied the output schema directly. The exact prompts, Skill text, decoding configuration, parser specification, and output records are included in the supplement. Because the matched pair differs only by the Skill, their aggregate output difference isolates the condition-level change associated with adding the full procedure; instruction-level effects require separate ablations.

\subsection{Evaluation Measures}

We report accuracy and metaphor-class precision, recall, and F1. Macro-F1, the unweighted mean of both class F1 scores, is the primary cross-dataset measure because class prevalence differs. The stability analysis considers four complementary summaries: the mean across CCIME and CMC, the lower of those two external scores, the signed difference between CMRE Test and the external mean, and the range across all three datasets. A small range is interpreted together with the performance floor, because uniformly low scores would also produce little variation. Fine-tuned values are means and sample SDs over three seeds; zero-shot values are deterministic point estimates.

\section{Results}

\begin{figure*}[t]
\centering
\includegraphics[width=0.96\textwidth]{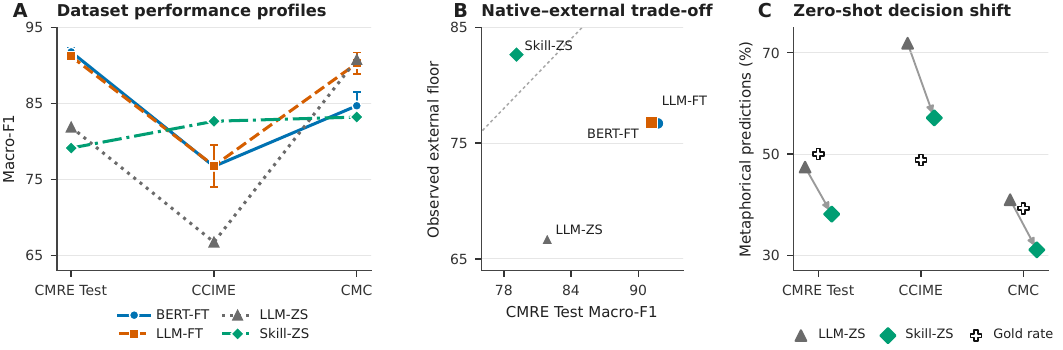}
\caption{Complementary profiles. (A) Macro-F1 by dataset. (B) CMRE Test Macro-F1 versus the observed external floor; the diagonal marks equality. (C) Rates of metaphorical predictions for the matched zero-shot pair and gold prevalence; arrows run from LLM-ZS to Skill-ZS. Whiskers show sample SD only for the three-seed fine-tuned conditions.}
\label{fig:profiles}
\end{figure*}

The analysis begins with the cross-dataset profile that motivates the comparison, then examines the class-specific trade-offs and matched decision shift behind it. Fig.~\ref{fig:profiles} visualizes the Macro-F1 and prediction profiles. Dataset scores are rounded to two decimals in Table~\ref{tab:main}; the stability criteria in Table~\ref{tab:transfer} use the corresponding unrounded values.

\subsection{Observed Cross-Dataset Stability}

\begin{table}[t]
\centering
\caption{Observed cross-dataset stability criteria in Macro-F1 points, computed from unrounded values. Native $-$ ext. mean is signed.}
\label{tab:transfer}
\footnotesize
\setlength{\tabcolsep}{2.7pt}
\begin{tabular}{lrrrr}
\toprule
Method & Ext. mean & Ext. floor & \begin{tabular}{@{}c@{}}Native $-$\\ext. mean\end{tabular} & Range \\
\midrule
BERT-FT & 80.69 & 76.70 & 11.07 & 15.06 \\
LLM-FT & \textbf{83.52} & 76.78 & 7.65 & 14.39 \\
LLM-ZS & 78.75 & 66.74 & 3.12 & 24.02 \\
Skill-ZS & 82.92 & \textbf{82.64} & -3.80 & \textbf{4.08} \\
\bottomrule
\end{tabular}
\end{table}

The main pattern is a separation between peak performance and cross-dataset evenness. BERT-FT has the highest native Macro-F1, and LLM-FT has the highest external mean at 83.52. Skill-ZS is only 0.60 points lower on the external mean, while attaining the highest external floor (82.64) and the smallest observed three-dataset range (4.08). Its floor exceeds those of LLM-FT and BERT-FT by 5.86 and 5.94 points, respectively. RQ1 therefore identifies Skill-ZS as having the most even observed profile among the four conditions, rather than the highest score on every criterion.

The signed native--external gap further distinguishes the profiles. Both fine-tuned models score higher on CMRE Test than on their external average, while Skill-ZS scores 3.80 points higher on average externally than natively. This direction is descriptive rather than intrinsically desirable; together, the external floor and range show that the Skill's observed stability is coupled with a lower native peak.

The different run designs limit the precision with which these point differences can be interpreted. BERT-FT and LLM-FT provide sample SDs across three seeds; the largest displayed dispersion is LLM-FT on CCIME at 2.77 points. The LLM-ZS and Skill-ZS results are deterministic point estimates under their frozen configurations and provide no comparable estimate of between-run dispersion. We therefore report rankings and differences descriptively, without significance claims for cross-regime contrasts. The figure reflects this distinction by drawing whiskers only for fine-tuned conditions.

\subsection{Dataset and Class-Specific Trade-offs}

\begin{table*}[t]
\centering
\caption{Classification performance (\%). P, R, and F1 refer to the metaphorical class. Fine-tuned values are means $\pm$ sample SDs over three seeds; zero-shot values are point estimates. Bold marks the highest observed value in each dataset--metric column.}
\label{tab:main}
\scriptsize
\setlength{\tabcolsep}{3.0pt}
\renewcommand{\arraystretch}{0.88}
\begin{tabular}{llccccc}
\toprule
Dataset & Method & Accuracy & P & R & F1 & Macro-F1 \\
\midrule
\multirow{4}{*}{CMRE Test}
 & BERT-FT & $\mathbf{91.76{\pm}0.51}$ & $\mathbf{92.70{\pm}0.16}$ & $90.67{\pm}1.30$ & $\mathbf{91.67{\pm}0.59}$ & $\mathbf{91.76{\pm}0.51}$ \\
 & LLM-FT & $91.18{\pm}0.51$ & $88.95{\pm}0.62$ & $\mathbf{94.04{\pm}0.59}$ & $91.42{\pm}0.49$ & $91.17{\pm}0.51$ \\
 & LLM-ZS & 81.88 & 83.62 & 79.29 & 81.40 & 81.87 \\
 & Skill-ZS & 79.41 & 88.58 & 67.53 & 76.64 & 79.12 \\
\midrule
\multirow{4}{*}{CCIME}
 & BERT-FT & $76.97{\pm}0.28$ & $70.85{\pm}0.67$ & $89.71{\pm}1.17$ & $79.17{\pm}0.12$ & $76.70{\pm}0.34$ \\
 & LLM-FT & $77.48{\pm}2.46$ & $69.35{\pm}2.57$ & $\mathbf{96.77{\pm}1.37}$ & $80.77{\pm}1.67$ & $76.78{\pm}2.77$ \\
 & LLM-ZS & 68.16 & 61.80 & 91.01 & 73.61 & 66.74 \\
 & Skill-ZS & \textbf{82.70} & \textbf{77.56} & 90.83 & \textbf{83.67} & \textbf{82.64} \\
\midrule
\multirow{4}{*}{CMC}
 & BERT-FT & $85.53{\pm}1.74$ & $83.44{\pm}3.06$ & $78.83{\pm}1.32$ & $81.06{\pm}2.10$ & $84.68{\pm}1.79$ \\
 & LLM-FT & $90.74{\pm}1.30$ & $\mathbf{88.86{\pm}1.24}$ & $87.35{\pm}3.04$ & $88.08{\pm}1.83$ & $90.25{\pm}1.41$ \\
 & LLM-ZS & \textbf{91.12} & 87.06 & \textbf{90.88} & \textbf{88.93} & \textbf{90.76} \\
 & Skill-ZS & 84.67 & 88.48 & 70.07 & 78.21 & 83.19 \\
\bottomrule
\end{tabular}
\end{table*}

On CMRE Test, BERT-FT and LLM-FT have the two highest Macro-F1 values but different class profiles: BERT-FT has higher metaphor precision (92.70 versus 88.95), whereas LLM-FT has higher recall (94.04 versus 90.67). Skill-ZS raises precision relative to LLM-ZS (88.58 versus 83.62) while lowering recall from 79.29 to 67.53.

On CCIME, Skill-ZS raises precision relative to LLM-ZS from 61.80 to 77.56 while maintaining similar recall (90.83 versus 91.01). Accuracy, metaphor F1, and Macro-F1 increase by 14.54, 10.06, and 15.90 points. LLM-FT has the highest recall but lower precision than BERT-FT, leaving their Macro-F1 values nearly equal.

On CMC, LLM-ZS has the highest accuracy, recall, metaphor F1, and Macro-F1, while LLM-FT has the highest precision. Skill-ZS raises precision slightly relative to LLM-ZS (88.48 versus 87.06) but lowers recall from 90.88 to 70.07. RQ2 therefore shows that the Skill's more even Macro-F1 profile emerges from dataset-specific precision--recall trade-offs rather than uniform gains in every measure.

Both fine-tuned models have their lowest Macro-F1 on CCIME. Skill-ZS instead scores 3.52 points higher on CCIME and 4.08 points higher on CMC than on CMRE Test. CCIME's treatment of similes and hyperbole as non-metaphorical offers one plausible explanation for this contrast, alongside other differences in dataset composition.

\subsection{Aggregate Decision Redistribution}

\begin{table*}[t]
\centering
\caption{Aggregate confusion counts and rates of metaphorical predictions for matched zero-shot conditions.}
\label{tab:confusion}
\footnotesize
\begin{tabular}{llrrrrrr}
\toprule
Dataset & Method & TN & FP & FN & TP & Predicted metaphorical & Gold metaphorical \\
\midrule
\multirow{2}{*}{CMRE Test}
 & LLM-ZS & 359 & 66 & 88 & 337 & 47.41\% & 50.00\% \\
 & Skill-ZS & 388 & 37 & 138 & 287 & 38.12\% & 50.00\% \\
\midrule
\multirow{2}{*}{CCIME}
 & LLM-ZS & 276 & 319 & 51 & 516 & 71.86\% & 48.80\% \\
 & Skill-ZS & 446 & 149 & 52 & 515 & 57.14\% & 48.80\% \\
\midrule
\multirow{2}{*}{CMC}
 & LLM-ZS & 387 & 37 & 25 & 249 & 40.97\% & 39.26\% \\
 & Skill-ZS & 399 & 25 & 82 & 192 & 31.09\% & 39.26\% \\
\bottomrule
\end{tabular}
\end{table*}

Table~\ref{tab:confusion} reports aggregate confusion counts and class-prediction rates for the matched zero-shot conditions. Skill-ZS predicts the metaphorical class less often on every dataset. On CCIME, its rate is 57.14\%, compared with 71.86\% for LLM-ZS; false positives fall from 319 to 149, while false negatives change from 51 to 52. On CMRE Test and CMC, false positives also fall, but false negatives rise by 50 and 57, respectively. RQ3 therefore identifies a consistent reduction in metaphorical predictions with dataset-dependent effects on error types.

Gold prevalence clarifies the contrast. On CCIME, LLM-ZS predicts 23.06 percentage points more metaphorical cases than the gold distribution, whereas Skill-ZS reduces that gap to 8.35 points. On CMRE Test and CMC, direct LLM-ZS is closer to gold prevalence than Skill-ZS. The same change in prediction tendency therefore improves alignment with the gold class prevalence on one dataset and worsens it on two others.

\section{Discussion}

The results reveal a trade-off between source-aligned peak performance and evenness across the observed datasets. We first interpret the Skill's cross-dataset profile, then relate that profile to its matched decision shift and to the choice between knowledge-supply regimes.

\subsection{A More Even Observed Cross-Dataset Profile}

The central finding concerns the shape of performance across datasets. Fine-tuning provides the strongest source-aligned results, and LLM-FT achieves the highest external average. Skill-ZS, by contrast, combines a competitive external mean with the highest observed external floor and the narrowest three-dataset range. Read together, the floor and range indicate that the Skill yields a more even performance level across the observed datasets, even though it does not maximize native performance.

This pattern extends cross-dataset evaluation beyond a single source-to-target drop. Prior work shows that genre, annotation consistency, and evaluation setting can alter metaphor-identification performance \cite{aghazadeh2022metaphors,uduehi2024expectation,reimann2024religious}. The present comparison distinguishes at least three desirable properties: a high native peak, a high external average, and a high external floor with limited variation. The explicit contextual- and basic-meaning criteria in the Skill may be less tied to CMRE-specific label regularities than supervised parameter updates, offering a plausible account of its more even profile. The results therefore support evaluating procedural guidance by both its level and its consistency across datasets.

\subsection{Stability Through a Decision Trade-off}

The matched zero-shot comparison shows how this profile is produced at the output level. Adding the Skill consistently lowers the rate of metaphorical predictions. On CCIME, where the direct model strongly overpredicts the positive class relative to gold prevalence, this shift removes many false positives with almost no recall loss. On CMRE Test and CMC, the same shift moves the predictions below the gold metaphor prevalence and produces substantially more false negatives. The Skill thus changes the operating tendency of the base model rather than delivering a uniform accuracy improvement.

Annotation policy provides a useful interpretation of this asymmetry. CCIME assigns similes and hyperbole to the non-metaphorical class, whereas CMRE retains released-positive similes. A stricter requirement for supported contextual--basic contrast can therefore align well with CCIME's negative-class boundary while suppressing true positives under another policy. Stability here refers to aggregate performance rather than invariant errors: the balance between precision and recall still changes by dataset. Similar operational criteria have been implemented through model structure, explicit sense comparison, prompting, and rule scripts \cite{mao2019sequential,choi2021melbert,elzohbi2024contrastwsd,reimann2025mipvu,huangliu2026rule}.

\subsection{Choosing a Knowledge-Supply Regime}

The comparison suggests that the preferred regime depends on the evaluation objective. Fine-tuning is favored when maximizing performance on the labeled source distribution is central, and LLM-FT offers the strongest external average in this study. Skill-ZS becomes attractive when the observed external floor and variation across datasets receive greater weight. Because its procedural criteria remain explicit at inference time, the Skill also supports direct revision and inspection without another parameter-update cycle.

These routes are complementary rather than mutually exclusive. One promising direction is to combine supervised adaptation with procedural checks that constrain overly broad metaphor predictions. Such hybrids should be assessed with class-specific measures, prespecified decision costs, controlled rule ablations, and prompt-sensitivity tests so that gains in average performance can be separated from gains in cross-dataset stability.

\section{Limitations}

The empirical scope contains four prespecified conditions, one base LLM, one expert-informed Skill, and three Chinese sentence-level datasets. CCIME and CMC provide two observed external settings; their external mean, external floor, and the three-dataset range do not establish behavior on unobserved genres, languages, annotation schemes, or metaphor constructions. Annotation policies are not fully aligned: CMRE retains source-positive similes as metaphorical, whereas CCIME assigns similes and hyperbole to its non-metaphorical class. Transfer to CCIME therefore partly measures adaptation to a changed labeling policy rather than domain change alone.

Fine-tuned conditions provide means and sample SDs across three seeds, whereas each zero-shot condition contributes one deterministic formal run. This asymmetry prevents comparable uncertainty estimates across all four conditions. The Skill was refined on 60 AI-assisted synthetic specification sentences, and the exact authoring-model identifier was not retained. The study did not independently reannotate those items or obtain external expert validation of every rule. Aggregate confusion counts identify the condition-level shift introduced by the complete Skill but cannot identify instruction-level causes; controlled ablation and prespecified instance-level analysis are needed for causal attribution.

The evaluation is sentence-level and binary. It neither identifies the exact metaphor-related word nor directly measures explanation quality, procedural compliance, calibration, or human interpretability. The two external datasets also differ in several properties, including resource construction and class distribution, so their performance differences cannot be attributed to a single domain-shift factor.

\section{Conclusion}

Across three Chinese metaphor datasets, the frozen expert-informed Skill produces the most even observed performance profile: it has the highest external floor and the smallest three-dataset Macro-F1 range, while its external mean remains within 0.60 points of LLM-FT. Fine-tuning retains the strongest native performance, and LLM-FT attains the highest external average. These outcomes distinguish cross-dataset stability from peak accuracy and show why both should be reported. To our knowledge, this is the first study to compare an expert-informed procedural Skill with task-specific fine-tuning in the same cross-dataset evaluation of Chinese sentence-level metaphor identification.

The matched zero-shot comparison connects the Skill's profile to a consistent reduction in metaphorical predictions. This shift substantially reduces false positives on CCIME while increasing false negatives on CMRE Test and CMC, demonstrating that a stable aggregate profile can contain different class-specific trade-offs. Expert-informed procedural prompting therefore provides a complementary route for supplying task knowledge: it yields a comparatively high performance floor across the observed datasets without updating model parameters. The profiles also motivate hybrid systems in which explicit operational criteria constrain training or post-training decisions while fine-tuning preserves source-specific sensitivity.

\section*{Ethics and Data Availability}

The study uses existing research datasets and model-generated predictions. No new personal or human-subject data were collected. The 60 synthetic specification sentences were produced through an AI-assisted authoring process for instruction development and are not presented as human annotation. Dataset access and redistribution remain subject to original release terms. The supplementary package will include preprocessing records, prompts, the frozen Skill, runtime configurations, instance-level predictions, evaluation code, and the figure source where redistribution terms permit.

\sloppy
\bibliographystyle{IEEEtran}
\bibliography{references_v11}

\begin{thebibliography}{10}
\providecommand{\url}[1]{#1}
\csname url@samestyle\endcsname
\providecommand{\newblock}{\relax}
\providecommand{\bibinfo}[2]{#2}
\providecommand{\BIBentrySTDinterwordspacing}{\spaceskip=0pt\relax}
\providecommand{\BIBentryALTinterwordstretchfactor}{4}
\providecommand{\BIBentryALTinterwordspacing}{\spaceskip=\fontdimen2\font plus
\BIBentryALTinterwordstretchfactor\fontdimen3\font minus
  \fontdimen4\font\relax}
\providecommand{\BIBforeignlanguage}[2]{{%
\expandafter\ifx\csname l@#1\endcsname\relax
\typeout{** WARNING: IEEEtran.bst: No hyphenation pattern has been}%
\typeout{** loaded for the language `#1'. Using the pattern for}%
\typeout{** the default language instead.}%
\else
\language=\csname l@#1\endcsname
\fi
#2}}
\providecommand{\BIBdecl}{\relax}
\BIBdecl

\bibitem{aghazadeh2022metaphors}
E.~Aghazadeh, M.~Fayyaz, and Y.~Yaghoobzadeh, ``Metaphors in pre-trained
  language models: Probing and generalization across datasets and languages,''
  in \emph{Proc. ACL}, 2022, pp. 2037--2050.

\bibitem{uduehi2024expectation}
O.~O. Uduehi and R.~C. Bunescu, ``An expectation-realization model for metaphor
  detection,'' in \emph{Proc. FigLang}, 2024, pp. 79--84.

\bibitem{reimann2024religious}
S.~Reimann and T.~Scheffler, ``Metaphors in online religious communication: A
  detailed dataset and cross-genre metaphor detection,'' in \emph{Proc.
  LREC-COLING}, 2024, pp. 11\,236--11\,246.

\bibitem{pragglejaz2007mip}
{Pragglejaz Group}, ``{MIP}: A method for identifying metaphorically used words
  in discourse,'' \emph{Metaphor and Symbol}, vol.~22, no.~1, pp. 1--39, 2007.

\bibitem{steen2010mipvu}
G.~J. Steen, A.~G. Dorst, J.~B. Herrmann, A.~A. Kaal, T.~Krennmayr, and
  T.~Pasma, \emph{A Method for Linguistic Metaphor Identification: From {MIP}
  to {MIPVU}}.\hskip 1em plus 0.5em minus 0.4em\relax John Benjamins, 2010.

\bibitem{mao2019sequential}
R.~Mao, C.~Lin, and F.~Guerin, ``End-to-end sequential metaphor identification
  inspired by linguistic theories,'' in \emph{Proc. ACL}, 2019, pp. 3888--3898.

\bibitem{choi2021melbert}
M.~Choi \emph{et~al.}, ``{MelBERT}: Metaphor detection via contextualized late
  interaction using metaphorical identification theories,'' in \emph{Proc.
  NAACL-HLT}, 2021, pp. 1763--1773.

\bibitem{li2023basic}
Y.~Li, S.~Wang, C.~Lin, and F.~Guerin, ``Metaphor detection via explicit basic
  meanings modelling,'' in \emph{Proc. ACL}, 2023, pp. 91--100.

\bibitem{reimann2025mipvu}
S.~Reimann and T.~Scheffler, ``Using large language models to perform
  {MIPVU}-inspired automatic metaphor detection,'' in \emph{Proc. Analogy-Angle
  II}, 2025, pp. 10--21.

\bibitem{huangliu2026rule}
W.~Huang and M.~Liu, ``Interpretable {Chinese} metaphor identification via
  {LLM}-assisted {MIPVU} rule script generation: A comparative protocol
  study,'' \emph{arXiv preprint arXiv:2603.10784}, 2026.

\bibitem{fuoli2026llm}
M.~Fuoli, W.~Huang, J.~Littlemore, S.~Turner, and E.~Wilding, ``Metaphor
  identification using large language models: A comparison of {RAG}, prompt
  engineering, and fine-tuning,'' \emph{Applied Corpus Linguistics}, vol.~6,
  no.~2, 2026, {Art. no. 100204}, doi: 10.1016/j.acorp.2026.100204.

\bibitem{elzohbi2024contrastwsd}
M.~Elzohbi and R.~Zhao, ``{ContrastWSD}: Enhancing metaphor detection with word
  sense disambiguation following the metaphor identification procedure,'' in
  \emph{Proc. LREC-COLING}, 2024, pp. 3907--3915.

\bibitem{chen2024reasonable}
P.~Chen, C.~Yang, and Q.~Huang, ``Merely judging metaphor is not enough:
  Research on reasonable metaphor detection,'' in \emph{Findings of EMNLP},
  2024, pp. 5850--5860.

\bibitem{tsvetkov2014crosslingual}
Y.~Tsvetkov, L.~Boytsov, A.~Gershman, E.~Nyberg, and C.~Dyer, ``Metaphor
  detection with cross-lingual model transfer,'' in \emph{Proc. ACL}, 2014, pp.
  248--258.

\bibitem{yang2023ood}
L.~Yang \emph{et~al.}, ``Out-of-distribution generalization in natural language
  processing: Past, present, and future,'' in \emph{Proc. EMNLP}, 2023, pp.
  4533--4559.

\bibitem{sanchezbayona2025surface}
E.~Sanchez-Bayona and R.~Agerri, ``Metaphor and large language models: When
  surface features matter more than deep understanding,'' in \emph{Findings of
  ACL}, 2025, pp. 17\,462--17\,477.

\bibitem{sanchezbayona2026meta4xnli}
E.~Sanchez-Bayona and R.~Agerri, ``{Meta4XNLI}: A cross-lingual parallel corpus
  for metaphor detection and interpretation,'' \emph{Computational
  Linguistics}, vol.~52, no.~1, pp. 191--235, 2026.

\bibitem{chen2023cmre}
G.~Chen \emph{et~al.}, ``{Chinese} metaphorical relation extraction,'' in
  \emph{Findings of EMNLP}, 2023, pp. 9085--9095.

\bibitem{ccime2026}
{BLCU ICALL}, ``{CCL2026} {Chinese} conversational meaning and metaphor
  evaluation,'' 2026, {[Online]. Available:}
  \url{https://github.com/blcuicall/CCIME2026}. {Accessed: Aug. 24, 2026}.

\bibitem{li2022cmgen}
Y.~Li, C.~Lin, and F.~Guerin, ``{CM-Gen}: A neural framework for {Chinese}
  metaphor generation with explicit context modelling,'' in \emph{Proc.
  COLING}, 2022, pp. 6468--6479.

\bibitem{cui2020chinese}
Y.~Cui, W.~Che, T.~Liu, B.~Qin, S.~Wang, and G.~Hu, ``Revisiting pre-trained
  models for {Chinese} natural language processing,'' in \emph{Findings of
  EMNLP}, 2020, pp. 657--668.

\bibitem{qwen2026qwen36}
{Qwen Team}, ``{Qwen3.6-27B} model card,'' 2026, {[Online]. Available:}
  \url{https://huggingface.co/Qwen/Qwen3.6-27B}. {Accessed: Aug. 24, 2026}.

\bibitem{hu2021lora}
E.~J. Hu \emph{et~al.}, ``{LoRA}: Low-rank adaptation of large language
  models,'' in \emph{Proc. ICLR}, 2022, {[Online]. Available:}
  \url{https://openreview.net/forum?id=nZeVKeeFYf9}. {Accessed: Aug. 24, 2026}.

\bibitem{dettmers2023qlora}
T.~Dettmers, A.~Pagnoni, A.~Holtzman, and L.~Zettlemoyer, ``{QLoRA}: Efficient
  finetuning of quantized {LLM}s,'' in \emph{Advances in Neural Information
  Processing Systems}, vol.~36, 2023, pp. 10\,088--10\,115.

\end{thebibliography}

\end{document}